\documentclass[runningheads]{llncs}
\usepackage[T1]{fontenc}
\usepackage{graphicx}
\usepackage{xcolor}
\usepackage{subfigure}
\usepackage{amssymb}
\begin{document}
\title{Effective black box adversarial attack with handcrafted kernels}
%
%
\author{Petr Dvo\v r\'a\v cek\orcidID{0000-0002-5525-7706} \and
Petr Hurtik\orcidID{0000-0003-4349-9705} \and
Petra \v Stevuli\'akov\'a\orcidID{0000-0002-7879-1397}}
\authorrunning{P. Dvo\v r\'a\v cek et al.}
%
\institute{University of Ostrava, Centre of Excellence IT4Innovations\\
Institute for Research and Applications of Fuzzy Modeling\\
30. dubna 22, Ostrava,  Czech Republic\\
\email{Petr.Dvoracek@osu.cz}\\
\email{Petr.Hurtik@osu.cz}\\
\email{Petra.Stevuliakova@osu.cz}}
\maketitle              
\begin{abstract}
We propose a new, simple framework for crafting adversarial examples for black box attacks. The idea is to simulate the substitution model with a non-trainable model compounded of just one layer of handcrafted convolutional kernels and then train the generator neural network to maximize the distance of the outputs for the original and generated adversarial image. We show that fooling the prediction of the first layer causes the whole network to be fooled and decreases its accuracy on adversarial inputs. Moreover, we do not train the neural network to obtain the first convolutional layer kernels, but we create them using the technique of F-transform. Therefore, our method is very time and resource effective.

\end{abstract}


\section{Introduction}
As the utilization of machine learning within the industry continues to rise, concerns emerge regarding its susceptibility to being hacked using adversarial machine learning techniques. The field of adversarial machine learning aims to trick machine learning models by providing misleading input (an adversarial example) and includes both the generation and detection of adversarial examples~\cite{szegedy2015explaining}. The \emph{adversarial example} is an input to machine learning models that an attacker has deliberately designed to cause the model to make a mistake. In particular, it is a corrupted version of a valid input, where the corruption is done by adding a perturbation of a small magnitude to it. The adversarial example aims to appear "normal" to humans but causes misclassification by the targeted machine learning model. An \emph{adversarial attack} is then a method to generate adversarial examples.

We distinguish two kinds of adversarial attacks: a white box attack and a black box attack. In the white box attack, the adversary has complete access to the target model, including the model architecture and parameters, allowing the attacker to use the model's gradient to produce the most effective adversarial examples~\cite{szegedy2013intriguing}. On the contrary, in the black box attack scenario, the attackers do not have full access to the model and can only observe the output of its prediction~\cite{papernot2016transferability,carlini2017towards}.

\begin{figure}[!h]
    \centering
    \includegraphics[width=\textwidth]{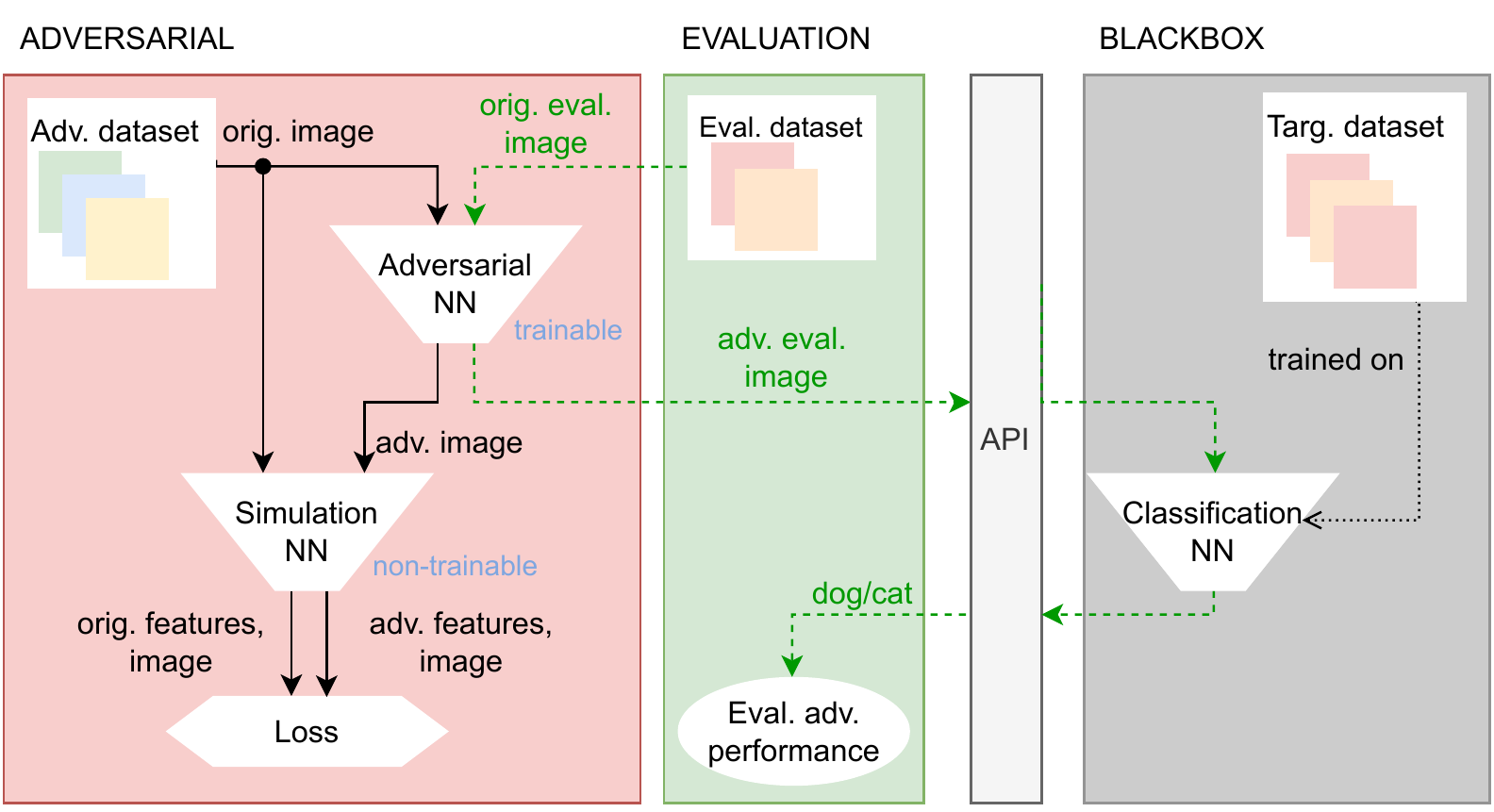}
    \caption{
    The proposed scheme. We consider three neural networks (NN): classification, simulation, and adversarial. The classification NN is a black box trained on an unknown 'target' dataset and presented by an unknown architecture. The simulation NN is not trainable and has one layer with 'simulation' kernels and serves for computing loss difference between the original and the adversarial output. The adversarial NN is a simple architecture, highly distinct from the black box one and trained on a different 'adversarial' dataset. Its goal is to create adversarial output that is similar to its input but maximizes the output of the simulation network. The trained adversarial NN is then used to distort images representing an input to the classification NN during the evaluation phase.
    }
    \label{fig:adver_training_diagram}
\end{figure}

Current methods to attack a neural network in a black box manner require training a custom substitution network on a dataset labeled with the probabilities from the output of the target network, which was acquired via an API (Application Programmable Interface)~\cite{papernot2016transferability}. For this purpose, custom images from the same domain as the target model must be acquired by the attacker and then labeled using many API requests to the model's API. Thousands of images, thus API requests are required to train the NN properly. This is not only inefficient but also brings unwanted attention to the attacker. A simple, time- and resource-efficient framework for performing a \textit{first-query} black box attack is presented in this paper. Our framework does not even require images from the same domain as the target model or training a huge classification neural network.

The framework we propose is shown in Figure~\ref{fig:adver_training_diagram} and consists of three main components: adversarial, evaluation, and black box. The main idea of this paper is captured in the adversarial component, which utilizes an adversarial dataset with classes different from the target dataset, an adversarial NN, a simulation NN, and a loss function. The adversarial NN is a trainable CNN generator of adversarial images. Simulation, on the other hand, is a non-trainable CNN with kernels predefined using the F-transform. The loss function consists of three parts: Mean Absolute Error (MAE) to minimize the global intensity difference of the produced image, Structural Similarity Index (SSIM) to force the network to produce an image visually similar to the original, and finally, the variability loss to prevent the trainable adversarial network from changing the color significantly. The framework itself is described in detail in Section~\ref{sec:exp}.

\section{Related work}

The first effective method of black box adversarial attack is described in~\cite{papernot2016transferability}. The trick is to substitute the targeted model with a custom-trained simulation model that can be fully controlled, allowing the attacker to obtain its gradients. To be able to train a simulation model similar to the target model, the attacker needs to create its own training dataset from the same domain as the model's domain. The labels for this dataset must be obtained as probability scores indicating the degree of association of the input with each class via the model's API.


The first published method to generate an adversarial example is \emph{L-BFGS} (Limited-memory BFGS)~\cite{Liu1989L-BFGS} which employs the \textit{Broyden–Fletcher–Goldfarb–\-Shanno} algorithm in the optimizer to create perturbations to the original image under $\ell_2$ norm. The BFGS-based optimizer is more resource-consuming but performs better in this task than the Adam or Adadelta optimizer. The creation of an adversarial example itself is slow in comparison with other methods, such as FGSM or BIM.

\emph{FGSM} (Fast Gradient Sign Method)~\cite{szegedy2015explaining} is much faster than the L-BFGS method mentioned for creating adversarial input. After obtaining the gradients from the neural network by forward and backward pass, the gradient sign in the image is taken, and the image itself is perturbed according to this sign by a small amount. The FGSM method itself is usually applied multiple times to increase its effectiveness. FGSM attack applied multiple times is called the BIM (Basic Iteration Method) attack~\cite{kurakin2016adversarial}.

The current framework for crafting adversarial inputs comes with several drawbacks:
\begin{itemize}
    \item It requires a pseudo-labeled dataset to train the simulation neural network.  To obtain this dataset, an excessive number of API queries must be sent to the target model, which could alert the victim about the ongoing attack and potentially lead to the discovery of the attacker.
    \item The training process of the simulation neural network can be time-consuming and computationally demanding, which could limit the feasibility of this attack in specific scenarios.
    \item Generating adversarial examples using this framework requires multiple forward or backward passes through the network, making execution on embedded devices or systems with limited computational power highly resource intensive and impractical.
\end{itemize}

We present a new framework for generating adversarial examples, which provides various benefits compared to the current methods. Our framework does not rely on data from the same domain and does not require many API calls to create pseudo-labels. Instead, adversarial examples are generated using a separate neural network that is significantly smaller in size than those used in current methods. This results in faster training times and lower resource requirements to generate adversarial inputs. Our approach offers a more efficient and practical solution for crafting adversarial examples that can be applied in various real-world scenarios.

\section{The proposed framework}
\begin{figure}[!h]
    \centering
    \subfigure[Mobilenet]{\includegraphics[width=.235\textwidth]{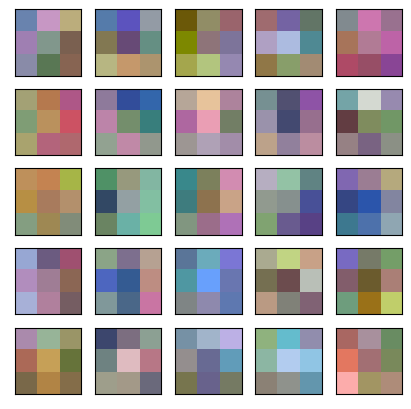}}
    \subfigure[NasNet]{\includegraphics[width=.235\textwidth]{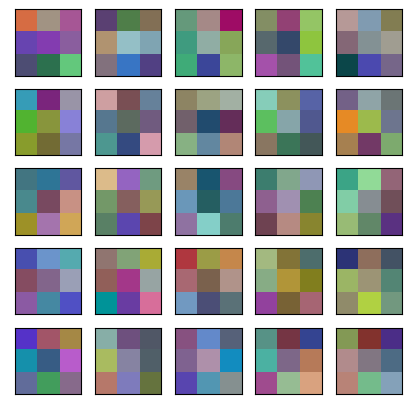}}
    \subfigure[ResNet50V2]{\includegraphics[width=.235\textwidth]{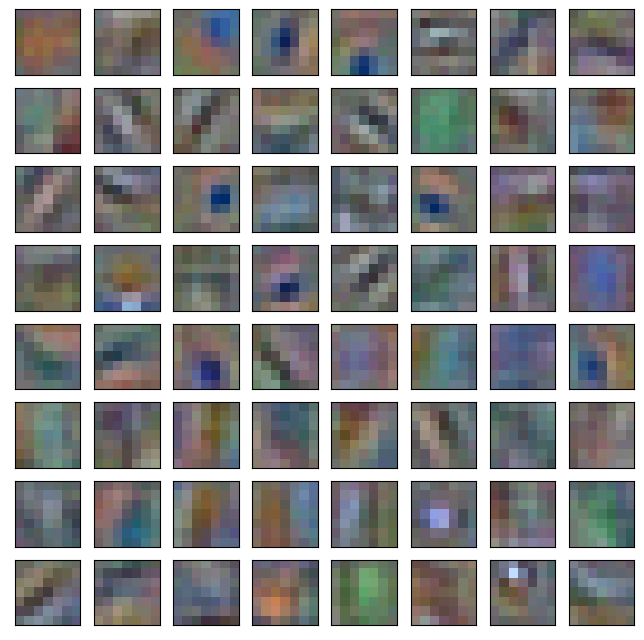}}
    \subfigure[DenseNet121]{\includegraphics[width=.235\textwidth]{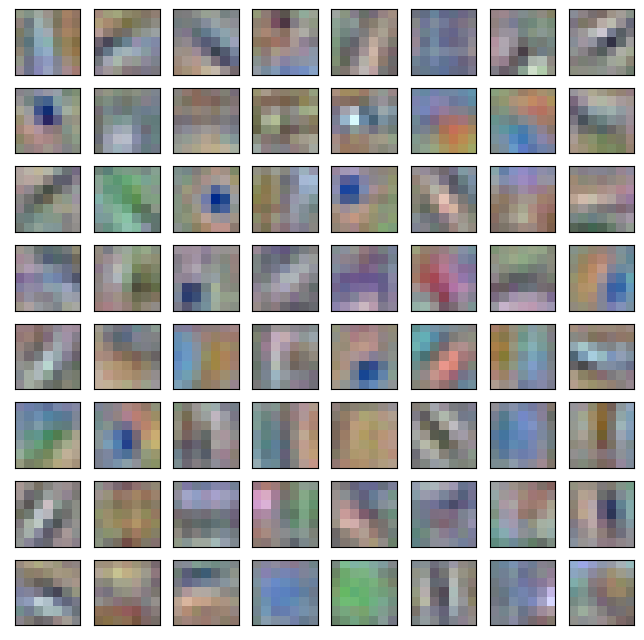}}\\
    ~\subfigure[CaffeeNet trained on ImageNet]{\includegraphics[width=.965\textwidth]{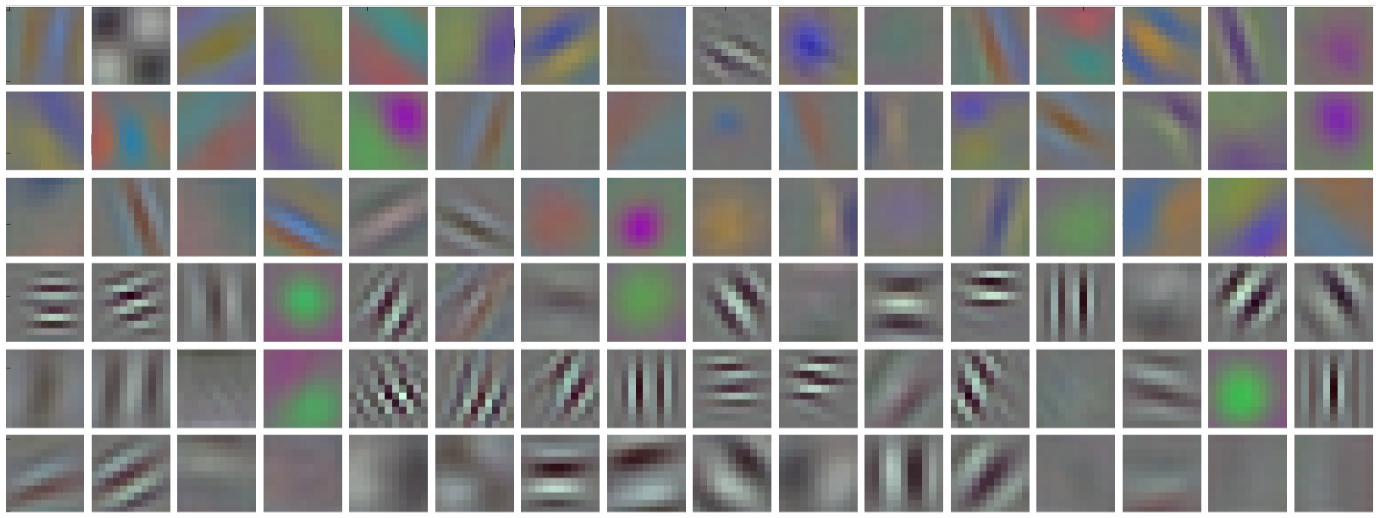}}
    \caption{Visualization of the first convolutional kernels in different neural network architectures. CaffeeNet model weights visualization was taken from~\cite{brachmann2016using}. The others are produced by neural networks used in this paper on CIFAR-10.}
    \label{fig:kernelsInNN}
\end{figure}

A feed-forward convolutional deep neural network sequentially maps the input to the desired output while starting with low-abstract features and consequently internally using more abstract features. It is known~\cite{bulat2020incremental} that low-abstract features are produced using kernels that are not task-dependent and serve as feature extractors for deeper layers. The feature representation of the input data should be complete in the sense of a possible reconstruction (backward representation) of any input object. Mathematically, this representation of an object is its approximation.

We aim to design a "simulation" convolutional neural network (CNN) that consists of universal kernels for extracting task-independent features representing input data. In general, most of the CNNs use in the first layer similar kernels that share the following characteristics: Gaussian-like, edge detection with various angle specifications, texture detection, and color spots; for example, see Figure~\ref{fig:kernelsInNN}. Therefore, kernels with such characteristics are usually those widely used in image processing, namely, Gaussian, Sobel, Kirsch, etc. Our idea of using universal kernels is based on kernels of possibly different dimensions and rotations. Unfortunately, standard image processing kernels lack such desired properties. Therefore, we propose to establish universal kernels based on the technique of F-transform (and F$^m$-transform)~\cite{PERFILIEVA2006FT,PerfDankBede:FT_Higher} which is a universal approximation technique already used in the field of CNNs~\cite{molek2020deep,molek2017convolutional}. More details about the proposed F-transform-based kernels are given in Section~\ref{sec:ftransform_kernels}.

\subsection{F-transform-based kernels}\label{sec:ftransform_kernels}
F-transform, in general, is a technique that transforms a function (image) into a component representation and back. Initially, the F-transform components were defined as constants representing the average values~\cite{PERFILIEVA2006FT}. Later, the F-transform of higher degree (F$^m$-transform) was defined~\cite{PerfDankBede:FT_Higher} where the F$^m$-transform components are represented by polynomials of degree $m$. The original F-transform is then the F$^0$-transform. In general, components can be understood as space-localized function features. The components are sufficient to reconstruct the original function with arbitrary precision. The process of generating components is dependent on the so-called basic functions (forming a fuzzy partition), their shape, width, and distance between them (for more details, see~\cite{PERFILIEVA2006FT}). The connection between the F-transform technique and convolution consists of the chosen basic functions; in particular, the basic function support corresponds to the width of the convolution kernel, the basic function shape corresponds to the shape of the convolution kernel, and the distance between the basic function nodes to the convolution stride.

The F$^m$-transform was already used to create convolutional kernels for CNN \cite{molek2020deep,molek2017convolutional}. The authors there presented that the shapes of the F$^m$-transform-based kernels are similar to the shapes of the kernels from the most well-known CNNs. In particular, the F$^0$-transform-based (FT$^0$) kernels are Gaussian-like, and the F$^1$-transform-based (FT$^1$) kernels are (horizontal or vertical) edge detectors that correspond to Sobel. In contrast to Sobel, the advantage is that the F$^m$-transform-based kernels are functionally defined and therefore can be easily expanded. Moreover, the F$^m$-transform-based kernels used in the first layer of CNN do not significantly change their shapes during training and are therefore an ideal choice for feature extraction.

\begin{figure}[!h]
    \centering
    \includegraphics[width=6mm]{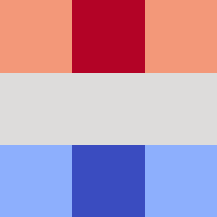}
    \includegraphics[width=10mm]{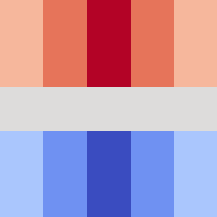}
    \includegraphics[width=14mm]{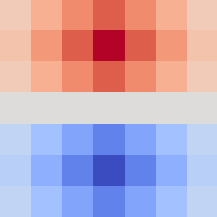}
    \includegraphics[width=18mm]{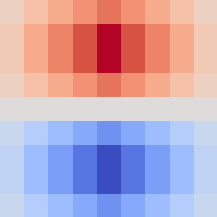}
    \caption{Visualization of simulation kernels represented by using FT$^1$. From left: kernel with dimension of 3$\times$3, 5$\times$5, 7$\times$7, and 9$\times$9.}
    \label{fig-visualization-of-kernels}
\end{figure}

In our simulation CNN, we particularly use F$^1$-transform-based kernels. We follow the theory and construction described in~\cite{molek2020deep} and compute the FT$^1$ kernels with dimension $d\times d$ (based on the triangular basic function of width $d$) as follows:
\begin{equation}
    FT^1(x,y) = \left(x-\frac{d+1}{2}\right)\left(1-\frac{|2x-d-1|}{d+1}\right)\left(1-\frac{|2y-d-1|}{d+1}\right),
\end{equation}
$$x,y \in [1, d].$$
We show the visualization of the kernels based on FT$^1$ with several dimensions in Figure~\ref{fig-visualization-of-kernels}.

In order to obtain kernels with edge detection features of various angles, we rotate the proposed FT$^1$ kernels by:
\begin{equation}
K_{\alpha} = K_x\cos \alpha + K_y\sin \alpha,
\end{equation}
where $K_x, K_y$ are kernels in the directions $x$ and $y$, respectively, and $K_{\alpha}$ is the desired kernel rotated with the angle $\alpha \in [0^\circ, 360^\circ]$. Visualization of an example of the $5\times 5$ FT$^1$ kernel rotated by $30^\circ$ is shown in Figure~\ref{fig-used-kernels}.

\section{Experimental verification}\label{sec:exp}

An overall visualization of the simulation framework to evaluate the effectiveness of the newly proposed black box adversarial attack is depicted in Figure~\ref{fig:adver_training_diagram}. It consists of three components: adversarial, black box, and evaluation. The black box component comprises the targeted classification NN model and its corresponding training data. As attackers, we are not granted direct access to the black box, but rather have the ability to invoke its API to obtain a prediction for a given input image. During the evaluation, data belonging to the model's domain (images of cats and dogs) are slightly modified by the adversarial generator to serve as the adversarial input. The targeted model's accuracy is assessed under this evaluation of adversarial attack. The adversarial component incorporates the adversarial image generator, implemented using a convolutional neural network, and a targeted neural network model simulator consisting of the F-transform-based kernels. The adversarial image generator is optimized through backpropagation to generate images that meet two criteria: 1) they should be similar to their original counterpart, and 2) when convolved with kernels from the simulator model, they should produce highly dissimilar vectors.

The CIFAR-10 dataset was used to represent the image classification task due to its low-performance requirements during training and testing. The attack was carried out against four common neural network architectures, namely ResNet-50 V2~\cite{he2016identity}, DenseNet 121~\cite{huang2017densely}, NASNetMobile~\cite{zoph2018learning} and MobileNet~\cite{howard2017mobilenets}.

\subsection{Data}
\begin{figure}[!h]
    \centering
    \includegraphics[width=14mm]{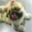}
    \includegraphics[width=14mm]{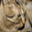}
    \includegraphics[width=14mm]{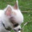}
    \includegraphics[width=14mm]{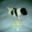}
    \includegraphics[width=14mm]{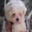}
    \includegraphics[width=14mm]{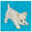}
    \includegraphics[width=14mm]{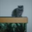}
    \includegraphics[width=14mm]{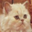}\\
    \includegraphics[width=14mm]{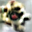}
    \includegraphics[width=14mm]{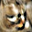}
    \includegraphics[width=14mm]{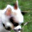}
    \includegraphics[width=14mm]{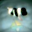}
    \includegraphics[width=14mm]{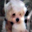}
    \includegraphics[width=14mm]{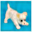}
    \includegraphics[width=14mm]{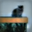}
    \includegraphics[width=14mm]{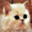}\\
    \includegraphics[width=14mm]{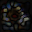}
    \includegraphics[width=14mm]{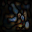}
    \includegraphics[width=14mm]{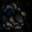}
    \includegraphics[width=14mm]{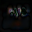}
    \includegraphics[width=14mm]{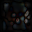}
    \includegraphics[width=14mm]{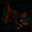}
    \includegraphics[width=14mm]{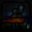}
    \includegraphics[width=14mm]{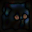}
    \caption{Samples from the classification's neural network test set. Top row: the original images. Middle row: images modified by the adversarial neural network. Bottom row: the absolute difference between the original and the adversarial image. The differences have a local manner and are connected to the main object in the image despite the adversarial model has been trained on different classes and without the information about the labels.}
    \label{fig-advers-images}
\end{figure}

We utilized CIFAR-10~\cite{krizhevsky2009learning} dataset with standard train/test split. From it, we extracted classes of cat and dog to be the aim of the black box neural network. It means that it is trained on 10,000 images and then tested on 4,000 images where these test images are fed into the network in their original and adversarial version. The remaining part of CIFAR, i.e. 40,000 training images and 6,000 test images, is used to train the adversarial network. The important thing is that the adversarial network does not have access during the training stage to the images that are devoted to the black box neural network. Only after training of the adversarial network is the test set of 4,000 images used to produce adversarial images just for the purpose of the attack performance evaluation. Image samples from the dataset can be seen in Figure~\ref{fig-advers-images}.

\subsection{Used neural networks}

\subsubsection{Classification neural network.} We involve a variety of neural networks (see Table~\ref{results-used-nn}) in order to evaluate the robustness of the proposed black box attack. We have selected neural networks that can process small input resolution of the dataset used, which is the reason why we have omitted EfficientNet~\cite{tan2019efficientnet} and ConvNeXt~\cite{liu2022convnet}.
\begin{table}[!ht]
\centering
\caption{Overview of the classification architectures used in the benchmark.}\label{results-used-nn}
\begin{tabular}{p{35mm}| p{25mm} p{25mm} p{25mm} }
\hline
Neural Network & Parameters & Depth & First kernels\\
\hline
ResNet-50 V2~\cite{he2016identity}& 25.6M &	103 &7$\times$7\\
DenseNet 121~\cite{huang2017densely} &8.1M	& 242 & 7$\times$7\\
NASNetMobile~\cite{zoph2018learning} & 5.3M &	389	& 3$\times$3\\
MobileNet~\cite{howard2017mobilenets} &4.3M & 55 & 3$\times$3\\
\hline
\end{tabular}
\end{table}
All networks are trained with Adadelta~\cite{zeiler2012adadelta} for 40k iterations, where the learning rate is 1.0 for the first 24iterations, 0.5 for the succeeding 10k iterations, and 0.1 for the last 6k iterations. The batch size is 64 and the loss function is binary crossentropy. As augmentation, we have chosen flips, perspective distortion, resize, rotate, and blur.

\subsubsection{Adversarial neural network.} As the objective of the adversarial network is to transform an image input into an image output, an encoder-decoder scheme such as U-Net~\cite{ronneberger2015u} complemented with a common backbone can be used. Because our dataset has a relatively small resolution, U-Net would map the input to too small feature maps. Therefore, we designed a straightforward architecture consisting of five convolutional layers without maxpooling/stride where each but the last convolutional layer is followed by batch normalization~\cite{ioffe2015batch} and ReLU activation. The last layer is connected to the sigmoid to obtain the output within the same range as the input image, that is, $[0, 1]$. It is obvious that the architecture used is very distinct from the architectures used for the classification neural network; from this point, it is the real black box adversarial attack.

\subsubsection{Simulation neural network.} 
\begin{figure}[!h]
    \centering
    \includegraphics[width=9.2mm]{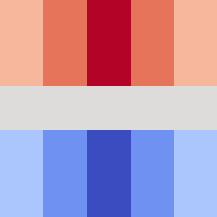}
    \includegraphics[width=9.2mm]{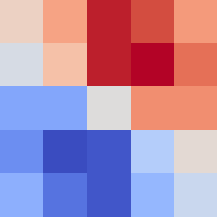}
    \includegraphics[width=9.2mm]{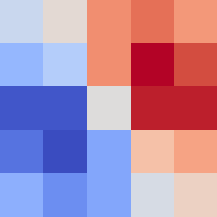}
    \includegraphics[width=9.2mm]{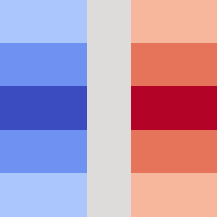}
    \includegraphics[width=9.2mm]{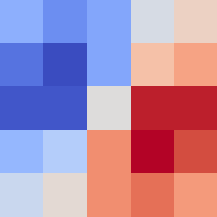}
    \includegraphics[width=9.2mm]{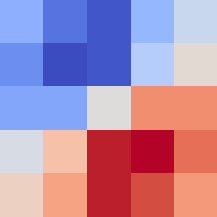}
    \includegraphics[width=9.2mm]{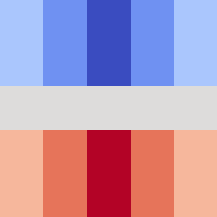}
    \includegraphics[width=9.2mm]{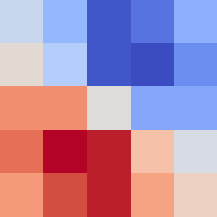}
    \includegraphics[width=9.2mm]{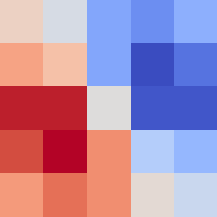}
    \includegraphics[width=9.2mm]{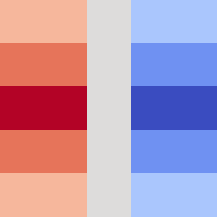}
    \includegraphics[width=9.2mm]{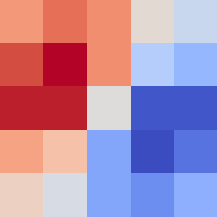}
    \includegraphics[width=9.2mm]{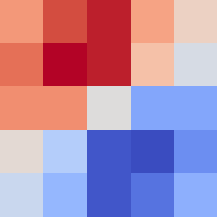}
    \caption{Visualization of the kernels' weights in the simulation neural network. We used $5\times 5$ FT$^1$ kernels rotated by $30^\circ$.}
    \label{fig-used-kernels}
\end{figure}

the network aims to produce a feature of the processed input image. It consists of $n$ depthwise convolutional layers that are connected into the input image (so that they process the same input on the same scale), and where each of the layers represents one particular FT$^1$ kernel. In our implementation, we use $5\times 5$ kernels that are mutually rotated by $30^\circ$ so that $n=12$. For visualization of the kernels, see Figure~\ref{fig-used-kernels}. The produced feature maps are finally concatenated. The whole network is not trainable to preserve the kernels in the predefined form.

\subsection{Loss function}
The loss function must consist of two parts, where the objective of the first is to produce the adversarial image visually similar to the original image, while the objective of the second part is to obtain a dissimilar feature mapping of these two images.
Formally, let $\psi$ be the simulation network consisting of FT kernels, $\omega$ be the adversarial network, and $I$ be the input image. Then, the compound loss function for training the adversarial network is defined as
$$
\ell(I) = \ell_S(I, \omega(I)) + | \psi(I) - \psi(\omega(I)) |,
$$
where $\ell_S$ is the similarity loss function given as 
$$
\ell_S(I_1,I_2) = |I_1-I_2|\alpha + \textrm{SSIM}(I_1, I_2)\beta +  \textrm{varc}(I_1, I_2)\gamma.
$$
Here, we use $\alpha, \beta, \gamma$ as weighting constants, SSIM be a structural similarity~\cite{wang2004image}, and varc a function that computes the variability in the channel dimension. It has been shown~\cite{ledig2017photo} that perceptual similarity suits human perception better than similarities derived from MSE, MAE, etc. Unfortunately, the perceptual similarity is based on neural networks, and there is a hypothesis that when our adversarial image confuses a general neural network, it will confuse the network that realizes similarity as well. That is the reason why we combined three partial statistic-based losses for similarity. The first part, MAE, minimizes global intensity difference, SSIM focuses on the texture, and variability of channels preserves colors.

\subsection{Numerical results}

All tested networks achieve nice results for the unmodified test set, but strongly fail for the test set modified by the adversarial network; see Table~\ref{results-used-nn}. In the case of the class 'dog', the models oscillate around the accuracy of 50\%, i.e., the accuracy of flipping the coin. On the other hand, the accuracy for class 'cat' increased, which is given by the fact that the models classify images (both cat and dog) more often as 'cat'. 

Visual observation of adversarial inputs visualized in Figure~\ref{fig-advers-images} led to the finding that adversarial inputs have a local increase in contrast. Therefore, we added augmentations of contrast, saturation, hue, and intensity into the training pipeline of the classification networks to observe how  robustness to adversarial attacks will be changed. Table~\ref{results-used-nn-aug} shows the results. 

\begin{table}[!ht]
\centering
\caption{Overview of the classification architectures used in the benchmark.}\label{results-used-nn}
\begin{tabular}{p{35mm}| p{3mm} p{15mm} p{15mm} p{15mm} p{15mm} }
\hline
& & \multicolumn{4}{c}{Accuracy on test set [\%]}\\
& & \multicolumn{2}{l}{Original input}& \multicolumn{2}{l}{Adversarial input}\\
Neural network & &cat & dog & cat & dog\\
\hline
ResNet-50 V2~\cite{he2016identity}& & 78.9& 83.9& 82.2& 50.1\\
DenseNet 121~\cite{huang2017densely} &&79.2& 82.7& 84.3& 57.7\\
NASNetMobile~\cite{zoph2018learning} && 74.8 & 86.6& 80.5 & 46.2 \\
MobileNet~\cite{howard2017mobilenets} && 77.1 & 85.0& 80.8& 49.6\\
\hline
\end{tabular}
\end{table}

\begin{table}[!ht]
\centering
\caption{Overview of the classification architectures used in the benchmark. The models were trained with enriched augmentations of contrast, saturation, hue, and intensity.}\label{results-used-nn-aug}
\begin{tabular}{p{35mm}| p{3mm} p{15mm} p{15mm} p{15mm} p{15mm} }
\hline
& & \multicolumn{4}{c}{Accuracy on test set [\%]}\\
& & \multicolumn{2}{l}{Original input}& \multicolumn{2}{l}{Adversarial input}\\
Neural network & &cat & dog & cat & dog\\
\hline
ResNet-50 V2~\cite{he2016identity}&& 81.7 & 85.6& 77.7 &53.5  \\
DenseNet 121~\cite{huang2017densely} &&81.2  &85.5 &84.5  &58.4  \\
NASNetMobile~\cite{zoph2018learning} &&  71.5& 83.5 & 80.5 & 52.7\\
MobileNet~\cite{howard2017mobilenets} &&79.4  &85.6 &81.0 &51.4 \\
\hline
\end{tabular}
\end{table}

\section{Conclusion}

We have presented the framework for generating adversarial examples in the black box manner that does not require access to any target model information, its training dataset, or the logit probabilities of its output via an exposed API. This was achieved by simulating the substitution model using the non-trainable model that employed only the single convolutional layer. The convolutional kernels in this layer were handcrafted using the F-transform to simulate the kernels that are usually created by the neural network without the need to train the substitution classification neural network itself. This method of simulating the substitution neural network instead of training seems to be promising, mainly because the need to access the target model logits is eliminated and the creation of the adversarial examples is more time and resource efficient. Verification of attack performance was performed using the CIFAR10 dataset. 

\subsubsection{Acknowledgements} This work was partially supported by SGS13/PřF-MF/2023.

%
%
%
\bibliographystyle{splncs04}
\bibliography{literature}

\end{document}